\pdfoutput=1

\documentclass[11pt]{article}

\usepackage[]{acl}

\usepackage{times}
\usepackage{latexsym}

\usepackage[T1]{fontenc}

\usepackage[utf8]{inputenc}

\usepackage{microtype}

\usepackage{xspace}
\usepackage{booktabs}
\usepackage{multirow}
\usepackage{hyperref}
\usepackage{url}
\usepackage[normalem]{ulem}
\usepackage{enumitem}
\usepackage{amsmath}
\usepackage{graphicx}
\usepackage{amssymb}
\usepackage{wrapfig}
\usepackage{subfig}
\usepackage{caption}
\usepackage{lineno}

\newcommand{\seedword}{\mbox{\textsc{Seed}}\xspace}
\newcommand{\prompt}{\mbox{\textsc{Prompt}}\xspace}
\newcommand{\xws}{\mbox{XWS}\xspace}
\newcommand{\tc}{\mbox{TC}\xspace}

\title{A Benchmark on Extremely Weakly Supervised Text Classification: Reconcile Seed Matching and Prompting Approaches}

\newcommand*\samethanks[1][\value{footnote}]{\footnotemark[#1]}

\author{Zihan Wang$^1$\thanks{$\ \ $Equal Contribution.} $\ \ $ Tianle Wang$^2$\samethanks $\ \ $ Dheeraj Mekala$^1$ $\ \ $Jingbo Shang$^1$\thanks{$\ \ $Corresponding Author.}\\
\small $^1$ University of California, San Diego \\
\small $^2$ Shanghai Jiao Tong University \\
  \{ziw224, dmekala, jshang\}@ucsd.edu $\ \ $ wtl666wtl@sjtu.edu.cn
}
\begin{document}
\maketitle
\begin{abstract}
E\underline{X}tremely Weakly Supervised Text Classification (\xws-\tc) refers to text classification based on minimal high-level human guidance, such as a few label-indicative seed words or classification instructions. 
There are two mainstream approaches for \xws-\tc, however, never being rigorously compared:
(1) training classifiers based on pseudo-labels generated by \textit{(softly) matching seed words} (\seedword) and
(2) \emph{prompting (and calibrating)} language models using classification instruction (and raw texts) to decode label words (\prompt). 
This paper presents the first \xws-\tc benchmark to compare the two approaches on fair grounds, where the datasets, supervisions, and hyperparameter choices are standardized across methods. 
Our benchmarking results suggest that
(1) Both \seedword and \prompt approaches are competitive and there is no clear winner; 
(2) \seedword is empirically more tolerant than \prompt to human guidance (e.g., seed words, classification instructions, and label words) changes;
(3) \seedword is empirically more selective than \prompt to the pre-trained language models; 
(4) Recent \seedword and \prompt methods have close connections and a clustering post-processing step based on raw in-domain texts is a strong performance booster to both. 
We hope this benchmark serves as a guideline in selecting \xws-\tc methods in different scenarios and stimulate interest in developing guidance- and model-robust \xws-\tc methods\footnote{Github repo at \url{https://github.com/ZihanWangKi/x-TC}}.

\end{abstract}

\section{Introduction}
\begin{figure}[t]
    \centering
    \includegraphics[width=\linewidth]{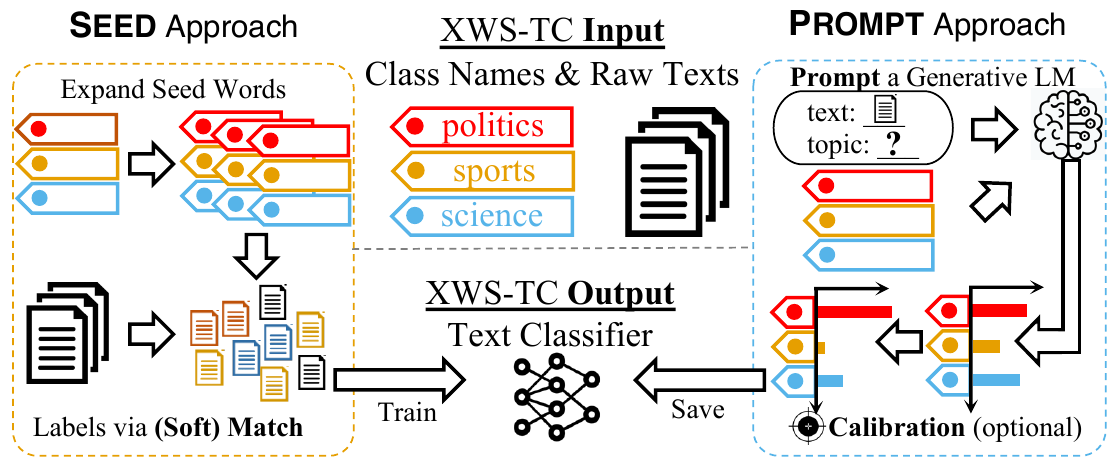}
    \caption{
    Illustrations of the XWS-TC problem and the \seedword and \prompt approaches.
    }
    \label{fig:method}
\end{figure}

Recently there has been a significant advancement in the text classification with the emergence of Extremely Weakly Supervised Text Classification (\xws-\tc) methods~\cite{lotclass, xclass, classkg, npprompt, lime}, which requires
no human-annotated datasets. Instead, these methods rely on minimal human guidance, such as the names of the classes or instructions describing the classification task. There are two main approaches to \xws-\tc: one based on matching seed words (\seedword), and the other on prompting a language model (LM) with instructions (\prompt). We give a brief introduction in the following paragraphs, and a more thorough review is in Section~\ref{sec:background}.

\seedword methods for \xws-\tc rely on a user-specified list of \textit{seed words} for each class, as well as an unlabeled in-domain corpus. These seed words are then expanded into a larger set of \textit{related words} for the class through statistical methods~\cite{conwea}, embedding similarity~\cite{xclass}, or masked language model predictions~\cite{lotclass}. These related words are used to assign a pseudo-class to each text in the unlabeled corpus through some matching strategy (e.g., assign a text to a class if it contains the related words for that class). The pseudo labels are then used to train a classifier through standard fully-supervised fine-tuning. 

On the other hand, \prompt methods for \xws-\tc, rely on reformulating text using an instruction template and prompting the language model to generate the likelihoods for each label in the classification task~\cite{gpt3}. 
For example, in a sentiment classification task, using an instruction template of \verb|<text>. sentiment:|, the model generating ``happy'' or ``sad'' will help classifiy the sentiment of the text.
Naive zero-shot prompting considers the highest likelihood label as the answer and recent improvements for more accurate likelihoods include calibration of likelihood scores~\cite{dcpmi,contextcal,protocal} and verbalizers that find more label words to better represent the class~\cite{PET, ProtoVerb, KPT}.

Both \seedword and \prompt methods have demonstrated strong performance in \xws-\tc.
However, there has been a lack of comprehensive comparison between these two approaches. 
This is due to the perception that the approaches are unrelated and the lack of standardization in datasets, supervision, and hyperparameter choices across methods.

We are motivated to construct a benchmark that fairly evaluates the performance of \xws-\tc methods. The benchmark consists of 11 datasets covering four domains along with their fine-grained variants and different numbers of classes. In addition, we make an effort to use the same hyperparameters across datasets for the methods, as there should not be a development set to tune the hyperparameters in the \xws setting~\cite{truefewshot}. 

Our benchmarking results suggest that both \seedword and \prompt approaches are competitive, with no clear winner. \seedword tends to perform better when both approaches use a similar-sized pre-trained model and is more robust and tolerant to changes in human guidance (such as seed words, classification instructions, and label words). On the other hand, \prompt methods have the ability to handle more general types of human guidance (such as descriptions of class names, rather than specific words) and do not have a strict requirement 
for an unlabeled corpus. When the underlying pre-trained language model changes, \prompt is more robust and scales better with the language model than \seedword. We also examine two specific methods from each approach, X-Class~\cite{xclass} and ProtoCal~\cite{protocal}, which independently proposed a  post-processing approach to calibrate the class predictions through clustering on an unlabeled in-domain corpus to improve classification performance. Our results show that this subroutine can be a universal booster for both \seedword and \prompt approaches.

Through this benchmark, we aim to advance the study of \xws-\tc methods and call for the development of methods that are robust to different human guidance and language models. 
We firmly believe that this paper will serve as a guide for selecting the appropriate method in different scenarios and contribute to the advancement of the field.

\section{Related Work}

\subsection{Different Types of Weak Supervision}
Extremely Weak Supervision is a setting that assumes access to only high-level human inputs, such as names of classes or instructions about classification criteria. 
We briefly discuss different types of minimal supervision in the following paragraphs.

\paragraph{Few-shot Supervision}
Few-shot supervision is the setting where there are only a small number of labeled examples for each of the classes. 
An intuitive way is to directly train the classifier on few-shot data, but usually that yields subpar performance.
Another popular way is called \textit{in-context learning}, where the few-shot supervision is used as \textit{context} to prompt LM for the answer~\cite{gpt3}.
Various methods have been proposed to improve it by
searching for better label words~\cite{PET, ProtoVerb}, stabilizing the output~\cite{promptorder}, and efficient fine-tuning~\cite{lmbff}.

\paragraph{Distant Supervision}
Distant supervision includes supervision from external resources such as encyclopedias or gazetteers.
There have been efforts to incorporate external knowledge into prompting~\cite{KPT}, phrase mining~\cite{autophrase}, and named entity recognition~\cite{bond}.
External models can also be used to help with extremely weak supervision. A line of research is on leveraging models trained on natural language inference data to suggest better-related words~\cite{lime} or directly classify the text~\cite{nli, self_nli}.

\paragraph{No Supervision}
Unsupervised methods fall into this category where they require no supervision.
These methods typically take one of the two following approaches: (1) clustering~\cite{unsup}, (2) topic modeling~\cite{lda}. 
However, both of these approaches lack control over the clusters/topics generated i.e. classes. 
For example, a text corpus can be categorized on several basis including topic, location, and sentiment. An unsupervised method cannot handle such scenarios. 
It would be beneficial to be able to retrieve all possible classifications of a corpus in an unsupervised manner, but as far as we are aware, there are no methods with this ability.

\subsection{Weak Supervision Benchmarks}
We introduce two other Weak Supervision Benchmarks and talk about differences with this work.

Wrench~\cite{wrench} is a benchmark that explored various types of weak supervision labeling functions (i.e., rules used to label the text). They synthesize the performance of different labeling functions, ways to combine them, and the fine-tuning process to learn the pseudo-training data. In our benchmark, we analyze extremely weak text classifiers that go beyond the labeling functions and compare their performance and robustness with zero-shot prompting.

AutoWS-Bench-101~\cite{AutoWS-Bench-101} is another benchmark that analyzes how labeling functions help text classification along with additional few-shot supervision.
They conclude that pre-trained models are strong baselines for in-domain settings and should be considered integrating with weak supervision methods. In this work, we focus on extremely weak supervision methods without any labeled data. The \seedword and \prompt methods compared in this benchmark are all based on pre-trained language models.

\subsection{Verbalizers}
Verbalizers are a type of \prompt method that find a larger set of label words so that the class choices are accurately represented.
We did not consider Verbalizer methods in this benchmark since they mostly rely on additional supervision, such as few-shot~\cite{PET,ProtoVerb} or an external knowledge base~\cite{KPT}. 









\section{Background}\label{sec:background}

Extremely Weak Supervision in Text Classification refers to a few high-level human guidance as supervision. This guidance typically is in the form of seed words that describe each class, or an instruction paired with label words that define the task. There are two main approaches for \xws-\tc: matching seed words (\seedword) and prompting language models (\prompt).

\subsection{Seed Matching Methods}

\seedword approaches are provided with a few class-indicative seed words and unlabeled documents as input. These methods typically involve seed word expansion where more words related to provided seed words are identified in the unlabeled corpus through several statistics-based~\cite{tfidf, conwea} or deep learning-based strategies~\cite{lotclass,xclass,classkg}.
Using these expanded seed words, each unlabeled document is pseudo-labeled. Different heuristics have been explored for pseudo-labeling such as string-matching~\cite{westclass}.
Recently, the matching approach has also evolved into softer manners such as embedding-based matching~\cite{xclass}, and graph-based matching~\cite{classkg}, that can address conflicts in a principled manner during pseudo-labeling. 


We introduce 4 strong-performing \seedword methods to include in our benchmark.

\noindent\textbf{LotClass}~\cite{lotclass} obtains related words through predicting masked tokens in a masked language modeling trained model~\cite{bert}, over an unlabelled corpus. They match the text to related words by fine-tuning a model to predict the related words given a text. 

\noindent\textbf{XClass}~\cite{xclass} obtains related words by finding words that have similar representations. 
They construct class-oriented representations for text.
and match the text to related words by representation similarity. 
They also showed that the performance can be improved significantly by matching based on clusters from text representations.

\noindent\textbf{ClassKG}~\cite{classkg} models the dependence of related words as an annotating problem on the keyword graph.

\noindent\textbf{NPPrompt}~\cite{npprompt} obtains related words 
through embedding similarity from a pre-trained LM.
The related words are used as label words to prompt a generative LM for predictions, which are then aggregated as the matching result. To some extent, NPPrompt belongs to an intersection of \prompt and \seedword methods.

\subsection{Prompt Methods}
Prompting language models is another approach to extremely weak supervision in text classification. This approach involves prompting a generative language model with an instructive text and extracting the \textit{likelihoods} of different label words.
This approach does not require an unlabeled in-domain corpus and can be used to predict text in an online fashion. 
However, language models have been known to be biased towards text sequences more common in pre-training data, leading to instability in zero-shot \& few-shot settings. Recently proposed post-processing methods~\cite{dcpmi, protocal} have attempted to address this by calibrating the predicted probabilities using estimates of the model's bias towards each verbalized label.
We describe 2 calibration methods.



\noindent\textbf{DC-PMI}~\cite{dcpmi} considers a null prompt to obtain the raw likelihoods of language model to predict each label. Then, for each text, they modify the likelihood of the predicted label by marginalizing the raw ones.

\noindent\textbf{ProtoCal}~\cite{protocal} considers an unlabelled corpus and obtains the predicted likelihoods on the corpus. 
The likelihood vectors are then clustered to better obtain the prediction boundary for each class. Instead of maximum likelihood, this prediction boundary is used to predict the class.

Some more \seedword and \prompt methods are described in Appendix~\ref{sec:appendix_other}.
\section{Benchmark}
\begin{table*}[t]
    \centering
    \renewcommand\tabcolsep{3pt}
    \begin{tabular}{ll ccccc}
    \toprule
    \textbf{Name} & \textbf{Domain}   & \# \textbf{Classes} & ||\textbf{Unlabelled}|| & ||\textbf{Eval}|| & \textbf{Imbalance} \\
    \midrule
    IMDB          & Reviews/Sentiment & 2                   & 5000                    & 5000              & 1.0 \\ 
    Yelp-2        & Reviews/Sentiment & 2                   & 5600                    & 3800              & 1.1 \\ 
    Yelp-5        & Reviews/Sentiment & 5                   & 6500                    & 5000              & 1.1 \\ 
    AGNews        & News/Topic        & 4                   & 6000                    & 7600              & 1.0 \\ 
    20News        & News/Topic        & 5                   & 6254                    & 5362              & 1.9 \\ 
    20News-Fine   & News/Topic        & 17                   & 5589                    & 4792              & 1.3 \\ 
    NYT-S         & News/Topic        & 5                   & 4578                    & 3925              & 17.1 \\ 
    NYT-S-Fine    & News/Topic        & 26                   & 4034                    & 3459              & 96.3 \\ 
    NYT           & News/Topic        & 9                   & 5119                    & 6400              & 30.7 \\ 
    NYT-Loc       & News/Location     & 10                   & 5119                    & 6400              & 17.1 \\ 
    DBpedia       & Wikipedia/Ontology& 14                   & 5600                    & 7000              & 1.3 \\ 
    \bottomrule
    \end{tabular}
    \caption{Dataset statistics in our benchmark.}
    \label{tab:dataset}
    \vspace{-1em}
\end{table*}
In order to establish a benchmark that can accurately evaluate various \xws-\tc methods, it is essential to consider a range of factors: Dataset choices, Instructions, Label words, Hyperparameter control, use of Pre-trained Language Models, Metrics and ensure their consistency across all experiments. We will discuss each of these factors in detail in the following sections.

\subsection{Dataset}
We consider datasets from prior evaluations~\cite{dcpmi,xclass,lotclass} 
that contain data from diverse domains.
To facilitate the evaluation process, the size of the evaluation set for each dataset has been controlled to a few thousand instances. 
Additionally, as many \xws-\tc methods require the use of an unlabelled in-domain corpus, a similar-sized sample has been sampled from the training split to serve this purpose, with the evaluation set and unlabelled corpus being disjoint. 
The datasets have been uniformly sampled without altering the distribution of labels, thus preserving the imbalance ratio, which is defined as the ratio between the size of the largest class and the smallest class. 
The statistics of the datasets are presented in Table~\ref{tab:dataset}. Details of the sources of the datasets are in Appendix~\ref{sec:appendix_data}.

\subsection{Instructions and Label/Seed Words}
To fairly compare \seedword and \prompt methods, we need to provide equal amounts of human supervision. That means, for \seedword methods, we should only allow a single word for each class, matching the amount used for label words. For instructions, we consider simple ones that hint at the classification criteria~\cite{dcpmi}. Details choices can be found in Appendix~\ref{sec:appendix_instructions}.

\subsection{Metrics}
For evaluation metrics, we consider the macro F$_1$ score on a dataset-by-dataset basis, which values each class within a dataset equally. To understand the performance of a method on all datasets, we employ two metrics: the average of the macro F$_1$ scores, and a ranking-based metric that combines the ranking of methods on each dataset to obtain a scale-prone value~\cite{rankscore}.

\subsection{Hyperparameters}
Another crucial aspect of the benchmark is the number of hyperparameters utilized by each method. 
In the context of extremely weak supervision, we argue that it is unrealistic to use different hyperparameters for different datasets, as doing so would necessitate the use of a separate development set, thereby defeating the purpose of using only high-level human supervision~\cite{truefewshot}. 
Therefore, we slightly tune the hyperparameters on one of the datasets to rule out failing scenarios and then stick with a single choice of hyperparameters throughout all datasets.
Under this hyperparameter enforcement, the ideal method should exhibit consistent performance across all datasets.

\subsection{Pre-trained Language Models}
\prompt methods use generative language models such as GPT while \seedword methods use representation encoding language models such as BERT. 
To fairly compare methods between these two approaches on \xws-\tc, we have to consider the ability of language models as a factor. 
We use the number of parameters of the pre-trained language model as an approximation of the power of the language model. 
Since all language models use the transformer as the backbone, this implies that the number of layers and size of hidden states is controlled. 
A further discussion is in Appendix~\ref{sec:appendix_plm}.

\subsection{Large Language Models}\label{sec:llm}
This benchmark specifically excludes the evaluation of (multi-task) fine-tuned language models such as T0~\cite{t0}, large language models (LLMs) such as GPT3, and human feedback-trained language models like Instruct-GPT~\cite{instructgpt} and ChatGPT because there are no equivalent representation encoding language models for the \seedword approaches. We discuss this in more details 
and include an evaluation of ChatGPT on a single dataset as a reference in Appendix~\ref{sec:appendix_llm}.

\begin{table*}[th]
    \centering
    \renewcommand\tabcolsep{3pt}
    \resizebox{\linewidth}{!}{
    \begin{tabular}{ll | ccc ccc ccc cc | cc}
    \toprule
    \textbf{Method} & \textbf{Model} & \textbf{IMDB} & \textbf{Yelp-2} & \textbf{Yelp-5} & \textbf{AGNews} & \textbf{20News} & \textbf{20News-Fine} & \textbf{NYT-S} & \textbf{NYT-S-Fine} & \textbf{NYT} & \textbf{NYT-Loc} & \textbf{DBpedia} & \textbf{Average} & \textbf{Rank Score} \\
    \midrule
    \multicolumn{15}{c}{\prompt} \\
    \midrule\midrule
    \multirow{2}{*}{Prompt}                             & GPT2-small   & 56.42 & 47.36 & 7.62 & 38.42 & 36.32 & 28.76 & 22.45 & 38.90 & 33.44 & 60.32 & 13.93 & 34.90 & 0 \\
    {}                                                  & GPT2-medium  & 35.80 & 33.57 & 25.87 & 69.36 & 55.16 & 46.03 & 54.08 & 46.14 & 24.92 & 79.00 & 24.52 & 44.95 & 1 \\
    \midrule
    \multirow{2}{*}{\shortstack[l]{Prompt\\+ DCPMI}}    & GPT2-small   & 70.13 & 65.34 & 23.01 & 72.67 & 61.64 & 37.45 & 73.93 & 63.19 & 55.20 & 70.40 & 51.10 & 58.55 & 4 \\
    {}                                                  & GPT2-medium  & 63.24 & 87.00 & 11.34 & 74.13 & 61.15 & 52.74 & 79.80 & 67.66 & 58.44 & 87.35 & 57.30 & 63.65 & 8 \\
    \midrule
    \multirow{2}{*}{\shortstack[l]{Prompt\\+ ProtoCal}} & GPT2-small   & 70.35 & 65.89 & 23.77 & 72.66 & 58.62 & 36.77 & 53.69 & 29.82 & 55.15 & 65.80 & 51.97 & 53.14 & 2 \\
    {}                                                  & GPT2-medium  & 70.58 & 88.60 & 36.62 & 75.26 & 62.58 & 48.55 & 51.97 & 46.85 & 59.04 & 72.45 & 66.46 & 61.54 & 9 \\
    \midrule
    \multicolumn{15}{c}{\seedword} \\
    \midrule\midrule
    \multirow{2}{*}{\shortstack[l]{LoT-Class}}          & BERT-base    & 58.56 & 67.96 & 24.92 & 73.94 & 70.57 & 9.40 & 61.36 & 23.05 & 48.59 & 67.13 & 57.98 & 51.2 & 3 \\
    {}                                                  & BERT-large   & 81.03 & 77.03 & 25.17 & 68.25 & 65.71 & 45.51 & 44.00 & 37.11 & 43.08 & 80.55 & 58.04 & 56.86 & 5 \\
    \midrule
    \multirow{2}{*}{\shortstack[l]{X-Class}}            & BERT-base    & 82.89 & 85.44 & 28.80 & 81.81 & 76.98 & 58.78 & 91.94 & 61.06 & 67.19 & 86.38 & 89.50 & 73.71 & 10 \\
    {}                                                  & BERT-large   & 82.05 & 90.39 & 31.02 & 85.91 & 77.52 & 59.98 & 87.53 & 68.40 & 68.73 & 85.77 & 87.91 & 75.02 & 12 \\
    \midrule
    \multirow{2}{*}{\shortstack[l]{ClassKG}}            & BERT-base    & 88.08 & 92.21 & 32.33 & 88.10 & 81.72 & 52.29 & 84.12 & 49.59 & 60.79 & 92.81 & 94.75 & 74.25 & 13 \\
    {}                                                  & BERT-large   & 90.96 & 93.10 & 39.41 & 87.30 & 83.84 & 51.62 & 80.95 & 59.95 & 56.31 & 91.03 & 72.74 & 73.38 & 11 \\
    \midrule
    \multirow{2}{*}{\shortstack[l]{NPPrompt}} & Roberta-base & 85.19 & 81.17 & 14.20 & 80.42 & 68.92 & 48.64 & 77.76 & 55.23 & 64.46 & 53.85 & 60.36 & 62.75 & 7 \\
    {}                                                  & Roberta-large& 85.67 & 93.58 & 23.45 & 83.62 & 69.82 & 43.33 & 77.93 & 35.91 & 59.96 & 65.83 & 47.11 & 62.38 & 6 \\
    \bottomrule
    \end{tabular}
    }
    \caption{Performance of \prompt and \seedword methods on the benchmark with standard models, prompt instructions, label words, and seed word choices. All scores are higher the better.}
    \label{tab:main}
\end{table*}

\section{Benchmark Experiments}
\begin{table*}[th]
    \centering
    \small
    \renewcommand\tabcolsep{3pt}
    \begin{tabular}{ll | ccc cccc | ccc}
    \toprule
    \multirow{2}{*}{\textbf{Method}} & \multirow{2}{*}{\textbf{Model}} & \multicolumn{7}{c}{\textbf{Yelp-2}} & \multicolumn{3}{|c}{\textbf{Averaged over Datasets}} \\
    & & \textbf{default} & \textbf{alt. 1} & \textbf{alt. 2} & \textbf{alt. 3} & \textbf{alt. 4} & \textbf{Median} & \textbf{Average (std)} & \textbf{Median} & \textbf{Average} & \textbf{std} \\
    \midrule
    \multicolumn{12}{c}{\prompt} \\
    \midrule\midrule
    \multirow{2}{*}{Prompt}                             & GPT2-small   & 47.36 & 49.34 & 32.84 & 58.19 & 32.24 & 47.36 & 43.99 (10.04) & 32.88 & 31.01 & 6.37 \\
    {}                                                  & GPT2-medium  & 33.57 & 32.89 & 32.84 & 55.10 & 32.78 & 32.89 & 37.44 (8.84) & 39.39 & 40.70 & 8.77 \\
    \midrule
    \multirow{2}{*}{\shortstack[l]{Prompt\\+ DCPMI}} & GPT2-small   & 65.34 & 57.19 & 72.80 & 45.12 & 56.98 & 57.19 & 59.49 (9.27) & 61.81 & 62.46 & 5.13 \\
    {}                                                  & GPT2-medium  & 87.00 & 66.65 & 36.53 & 75.31 & 39.23 & 66.65 & 60.94 (19.93) & 68.56 & 66.54 & 7.26 \\
    \midrule
    \multirow{2}{*}{\shortstack[l]{Prompt\\+ ProtoCal}} & GPT2-small   & 65.89 & 54.59 & 70.43 & 58.03 & 63.72 & 63.72 & 62.53 (5.63) & 64.62 & 64.03 & 6.17 \\
    {}                                                  & GPT2-medium  & 88.60 & 87.31 & 90.53 & 80.53 & 68.59 & 87.21 & 83.11 (8.00) & 72.17 & 70.74 & 8.76 \\
    \midrule
    \multicolumn{12}{c}{\seedword} \\
    \midrule\midrule
    \multirow{2}{*}{\shortstack[l]{X-Class}}            & BERT-base    & 85.44 & 88.01 & 85.69 & 62.24 & 84.33 & 85.44 & 81.14 (9.53) & 86.18 & 83.83 & 5.70 \\
    {}                                                  & BERT-large   & 90.39 & 89.71 & 88.70 & 84.75 & 85.49 & 88.70 & 87.81 (2.27) & 83.77 & 83.36 & 4.47 \\
    \midrule
    \multirow{2}{*}{\shortstack[l]{ClassKG}}            & BERT-base    & 92.21 & 91.71 & 87.78 & 91.18 & 92.47 & 91.71 & 91.07 (1.70) & 87.71 & 85.88 & 4.45 \\
    {}                                                  & BERT-large   & 93.10 & 93.16 & 94.13 & 93.89 & 92.01 & 93.16 & 93.26 (0.74) & 84.93 & 85.40 & 3.74 \\
    \bottomrule
    \end{tabular}
    \caption{Performance of \prompt and \seedword methods when the label word/seed word are changed to similar meaning alternatives. We show the performance on 5 choices of label words on Yelp-2 (4 alternatives + 1 default), its median, average, and standard deviation, and the averaged metrics across all datasets. }
    \label{tab:different_labels}
    \vspace{-1em}
\end{table*}
\begin{table*}[th]
    \small
    \centering
    \renewcommand\tabcolsep{3pt}
    \begin{tabular}{ll | ccc cccc | ccc}
    \toprule
    \multirow{2}{*}{\textbf{Method}} & \multirow{2}{*}{\textbf{Model}} & \multicolumn{7}{c}{\textbf{Yelp-2}} & \multicolumn{3}{|c}{\textbf{Averaged over Datasets}} \\
    & & \textbf{default} & \textbf{alt. 1} & \textbf{alt. 2} & \textbf{alt. 3} & \textbf{alt. 4} & \textbf{Median} & \textbf{Average (std)} & \textbf{Median} & \textbf{Average} & \textbf{std} \\
    \midrule
    \multirow{2}{*}{Prompt}                             & GPT2-small   & 47.36 & 32.89 & 37.31 & 73.11 & 39.01 & 39.01 & 45.94 (14.37) & 31.06 & 32.32 & 8.40 \\
    {}                                                  & GPT2-medium  & 33.57 & 33.18 & 56.77 & 78.41 & 42.34 & 42.34 & 48.85 (17.08) & 38.34 & 39.11 & 11.73 \\
    \midrule
    \multirow{2}{*}{\shortstack[l]{Prompt\\+ DMCPMI}} & GPT2-small   & 65.34 & 76.96 & 50.14 & 48.83 & 39.53 & 50.14 & 56.16 (13.29) & 60.00 & 61.48 & 6.45 \\
    {}                                                  & GPT2-medium  & 87.00 & 88.03 & 48.56 & 79.67 & 67.76 & 79.67 & 74.20 (14.72) & 65.26 & 61.54 & 14.18 \\
    \midrule
    \multirow{2}{*}{\shortstack[l]{Prompt\\+ ProtoCal}} & GPT2-small   & 65.89 & 83.87 & 60.54 & 71.23 & 72.25 & 72.25 & 70.76 (7.78) & 65.54 & 64.80 & 6.23 \\
    {}                                                  & GPT2-medium  & 88.60 & 87.40 & 57.85 & 80.13 & 82.73 & 82.73 & 79.34 (11.18) & 62.59 & 62.07 & 10.85 \\
    \bottomrule
    \end{tabular}
    \caption{Performance of \prompt methods when the instructions are changed to similar meaning alternatives. We show the performance on 5 choices of instructions on Yelp-2 (4 alternatives + 1 default), its median, average, and standard deviation, and the averaged metrics across all datasets. }
    \label{tab:different_instructions}

\end{table*}
\begin{table}[th!]
    \small
    \centering
    \resizebox{\linewidth}{!}{
    \begin{tabular}{ll | cc}
    \toprule
    \textbf{Method} & \textbf{Model} & \textbf{Average} & \textbf{Rank Score} \\
    \midrule
    \multicolumn{4}{c}{\prompt} \\
    \midrule\midrule
    \multirow{6}{*}{Prompt}                             & GPT2-small   & 30.54 & 1 \\
    {}                                                  & GPT2-medium  & 45.38 & 8 \\
    \cmidrule(lr){2-4}
    {}                                                  & BERT-base    & 43.04 & 7 \\
    {}                                                  & BERT-large   & 51.84 & 15 \\
    \cmidrule(lr){2-4}
    {}                                                  & RoBERTa-base & 45.71 & 6 \\
    {}                                                  & RoBERTa-large & 59.85 & 22 \\
    \midrule
    \multirow{6}{*}{\shortstack[l]{Prompt\\+ DCPMI}}    & GPT2-small   & 65.76 & 24 \\
    {}                                                  & GPT2-medium  & 74.56 & 31 \\
    \cmidrule(lr){2-4}
    {}                                                  & BERT-base    & 60.52 & 23 \\
    {}                                                  & BERT-large   & 55.88 & 14 \\
    \cmidrule(lr){2-4}
    {}                                                  & RoBERTa-base & 47.14 & 5\\
    {}                                                  & RoBERTa-large& 55.86 & 18 \\
    \midrule
    \multirow{6}{*}{\shortstack[l]{Prompt\\+ ProtoCal}} & GPT2-small   & 61.05 & 21 \\
    {}                                                  & GPT2-medium  & 70.07 & 30 \\
    \cmidrule(lr){2-4}
    {}                                                  & BERT-base    & 55.74 & 11 \\
    {}                                                  & BERT-large   & 70.16 & 25 \\
    \cmidrule(lr){2-4}
    {}                                                  & RoBERTa-base & 61.07 & 20 \\
    {}                                                  & RoBERTa-large& 66.09 & 28 \\
    \midrule
    \multicolumn{4}{c}{\seedword} \\
    \midrule\midrule
    \multirow{4}{*}{\shortstack[l]{X-Class}}            & BERT-base    & 87.17 & 37 \\
    {}                                                  & BERT-large   & 87.94 & 39 \\
    \cmidrule(lr){2-4}
    {}                                                  & RoBERTa-base & 60.18 & 19 \\
    {}                                                  & RoBERTa-large& 46.78 & 13 \\
    \midrule
    \multirow{4}{*}{\shortstack[l]{ClassKG}}            & BERT-base    & 89.80 & 40 \\
    {}                                                  & BERT-large   & 83.52 & 38 \\
    \cmidrule(lr){2-4}
    {}                                                  & RoBERTa-base & 86.94 & 36 \\
    {}                                                  & RoBERTa-large& 93.17 & 41 \\
    \midrule
    \multirow{4}{*}{\shortstack[l]{NPPrompt}}           & BERT-base    & 32.46 & 0 \\
    {}                                                  & BERT-large   & 31.45 & 2 \\
    \cmidrule(lr){2-4}
    {}                                                  & RoBERTa-base & 74.93 & 32 \\
    {}                                                  & RoBERTa-large& 75.56 & 33 \\
    \bottomrule
    \end{tabular}
    }
    \caption{Performance of \prompt and \seedword methods when the choice of the pre-trained model is alternated. }
    \label{tab:different_models}
    \vspace{-1em}
\end{table}

\subsection{Main Results}

In Table~\ref{tab:main} we show the performances of all \seedword and \prompt methods considered in the benchmark across the 11 datasets and report the average macro F$_1$ performance and the rank score. 

\paragraph{Performance of \prompt Methods}
We note that the performance of the standalone \prompt method is about 20 points lower than its counterparts with calibration methods. The use of additional instance independent instructions (DCPMI) or an additional clustering based on unlabelled text (ProtoCal) is crucial for \prompt methods to work well in \xws (zero-shot) text classification. 
\paragraph{Performance of \seedword Methods}
All the \seedword methods exhibit strong performance, with X-Class performing stably well across all datasets, and ClassKG performing the best on several datasets, but losing on certain fine-grained datasets. 

\paragraph{Comparing \prompt and \seedword Methods}
First, on the absolute performances, we can see that \seedword methods have overall better performance than \prompt methods, even when appropriate calibration is added for \prompt methods. However, we can also observe that a larger pre-trained GPT model increases the performance of \prompt methods quite significantly, while \seedword methods have a lower performance improvement when a larger pre-trained language model is used. This effect is further studied in Section~\ref{sec:exp_model}.

\subsection{Robustness}
Through this benchmark, we hope to not only decide which method performs the best, but also analyze under dynamic circumstances, which method is more robust to changes. 
Different choices of label words/seed words, instructions, and pre-trained language models can happen in real life.
Therefore, the robustness of methods when these ingredients are reasonably varied would indicate how stable the method is under variating circumstances. 
Due to the complexity of multiple runs of each method, we focus on 4 datasets pertaining to different domains, imbalance ratios, and number of classes: Yelp, AGNews, NYT-S, and DBpedia. We leave out two methods, LoT-Class and NPPrompt to save computational resources.

\subsubsection{Different Seed/Label words}\label{sec:exp_seed}
In Table~\ref{tab:different_labels} we explore the effect when a different choice of label words and seed words are used. For example, for Yelp-2, we chose negative/positive, terrible/great	bad/good, awful/find, and nasty/nice as the variants. We report the performance of the methods on each of the five choices, and also the aggregated performance over the 4 aforementioned datasets. We notice that \prompt methods in general have a high instability. 
While DCPMI and ProtoCal can remedy the variance a bit, \seedword methods are still more robust to changes of seed words. 

\subsubsection{Different Instructions}\label{sec:exp_instruction}
A high variance is also observed when the instructions are changed for the \prompt methods, as in Table~\ref{tab:different_instructions}. A noticeable trend is that when the pre-trained model is larger, while the performance increases, the variance brought by instructions or label words also increases. This could be alarming for \prompt methods.

\subsubsection{Different Pre-trained Language Models}\label{sec:exp_model}
In Table~\ref{tab:different_models} we analyze how changes in pre-trained language models would affect the performance of \seedword and \prompt methods (See Appendix~\ref{sec:appendix_differet_models_full} for the full table). 
Although \seedword performs better than \prompt,
\prompt methods has a strong increasing trend as the size of the pre-trained language model (e.g., changing from BERT-base to BERT-large). Also, X-Class and NPPrompt fail on RoBERTa and BERT respectively, which we hypothesize is that assumptions made in the methods are not general to all pre-trained language models; for example, the distribution of similarities of representations generated by a language model might be different by models.
This scaling trend is a factor that should be taken into selecting methods to use for \xws-\tc, when the language model size is different than evaluated in this benchmark.




\section{Connections between Recent \seedword and \prompt Methods}

\begin{figure*}
    \centering
    \includegraphics[width=\linewidth]{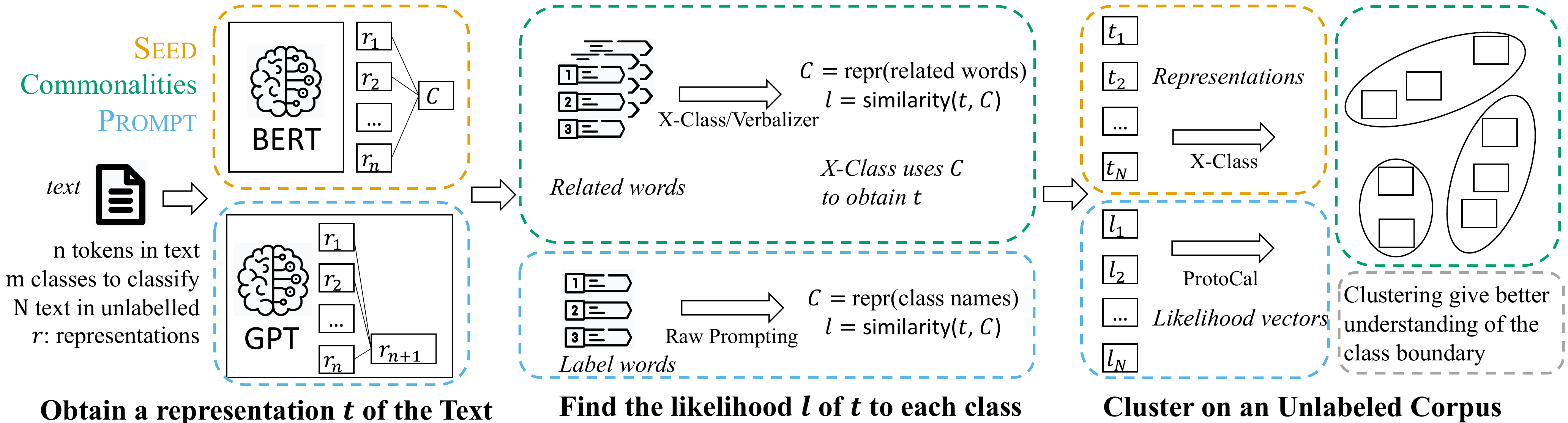}
    \caption{We highlight similarities (green) between a \seedword method X-Class (orange) and two \prompt methods Verbalizers and ProtoCal (blue). 
    }
    \label{fig:similarity}
    \vspace{-1em}
\end{figure*}

While \prompt is introduced by the seminal GPT-3 paper~\cite{gpt3} not too long ago, \seedword has a longer history and can be traced back to early tf-idf retrieval methods~\cite{tfidf}.
In recent years,  \seedword methods and \prompt methods are exploring similar ideas.
\seedword methods have been leveraging pre-trained language models to better understand the semantics of seed words; for example, by asking the language model to fill in masks~\cite{lotclass} or through means of representation similarities~\cite{xclass,npprompt}.
\prompt methods have been exploring calibration and verbalizers to improve and stabilize its predictions. Verbalizer includes a step of finding more label words that better represent the class, which is a similar approach used in \seedword.
We show that a recent representative \seedword method X-Class and two \prompt methods, Verbalizers and ProtoCal have higher similarities and deeper connections in their design.
This is particularly interesting as both directions have been developing independently.
In Figure~\ref{fig:similarity}, we provide a pipeline of the methods and highlight the similarities.

\subsection{Obtaining Text Representations}
X-Class matches text to classes by learning class-oriented text representations from an encoder-based language model.
X-Class views class representations as the union of representations describing the words.
The text representation in X-Class is defined as a weighted average of individual token representations where the weights are based on their respective similarity to the class representations.
On the other hand, general prompting relies on a decoder-based language model to produce a next token representation. In the penultimate layer of the decoder, the last token representation is computed by an attention mechanism over all other tokens,
which essentially produces a weighted average of all the token representations. 

In both methods, the text representation is obtained using an attention-like weighted average of tokens in the text.
The attention is guided such that the output representation is indicative of the class.
X-Class uses signals from class names to guide the attention while prompting relies on the understanding of the instruction.

\subsection{Obtaining Predicted Likelihoods}
\prompt methods obtain likelihoods of the class
by comparing the similarity of the next token representation to representations of the label words. A recent line of research on improving prompting for classification is to enlarge the set of label words to capture more diverse meanings of the classes, known as verbalizers, such as PET~\cite{PET}, ProtoVerb~\cite{ProtoVerb}, and KPT~\cite{PET,ProtoVerb,KPT}. 
The notion of verbalizers is very similar to seed-words expansion in \seedword methods. 
For example, X-Class and verbalizers 
both obtain a list of related words and use it to aggregate a class representation to replace the naive usage of label/seed word representation. Notably, the verbalizer methods require external supervision to find the related words, such as few-shot data~\cite{PET,ProtoVerb} or a knowledge base~\cite{KPT}
to obtain the related word list, while \seedword methods detect related words through an unlabelled corpus. 
Both approaches could be useful under different input settings.

\subsection{Unlabeled Corpus Clustering}\label{sec:cluster}
\begin{table}[t]
    \centering
    \renewcommand\tabcolsep{3pt}
    \resizebox{\linewidth}{!}{
    \begin{tabular}{ll | cc}
    \toprule
    \textbf{Method} & \textbf{Model} & \textbf{Average} & \textbf{Rank Score} \\
    \midrule
    Prompt                             & GPT2-small   & 34.90 & 0 \\
    Prompt + clustering                & GPT2-small   & 53.14 & 1 \\
    \midrule
    Prompt + DCPMI                            & GPT2-small   & 58.55 & 2 \\
    Prompt + + DCPMI + clustering                & GPT2-small   & 59.70 & 3 \\
    \midrule
    XClass (w/o clustering)            & BERT-base    & 67.40 & 6 \\
    XClass (w clustering)              & BERT-base    & 73.71 & 8 \\
    \midrule
    NPPrompt                           & roberta-base & 62.75 & 4 \\
    NPPrompt + clustering              & roberta-base & 64.54 & 5 \\
    \midrule
    ClassKG                            & BERT-base    & 74.25 & 7 \\
    ClassKG + clustering               & BERT-base    & 75.16 & 9 \\
    \bottomrule
    \end{tabular}
    }
    \caption{Performance of \prompt and \seedword methods with and without the clustering post-processing. }
    \label{tab:clustering}
\end{table}
Finally, a \seedword method X-Class and a \prompt method ProtoCal independently introduced a post-processing step by clustering on an unlabelled corpus, with the goal of obtaining a better decision boundary. 
X-Class clusters the text representations and initializes the clusters with the prior text-class similarity so that the clusters and classes are aligned. Protocal clusters the predicted likelihoods and align the clusters to classes by post-matching the cluster centers to the classes.
We further explore the effect of the two clustering ideas, a summary is in Table~\ref{tab:clustering} (Full table in Appendix~\ref{sec:appendix_clustering}). We show that adding such a post-clustering process to various methods can almost freely (apart from an unlabeled corpus) improve the performance of different methods consistently for five different methods.

\subsection{Implications}\label{sec:implications}
Given these connections between \seedword and \prompt methods and previous analysis on robustness, a natural extension is to analyze the cause of
the stability issues on label/seed words and model differences. 
We presented one empirical analysis of the clustering step in X-Class and ProtoCal and show that this step can improve performance for various different methods talked about in the benchmark (Section~\ref{sec:cluster}). Further analysis on other components is left as future work.
For example, one could reason that the introduction of related words makes the model less sensitive to the given label/seed words. 
This would require an exploration of the quality of the related words found by different \seedword and verbalizer methods, and whether the related words between methods can be used interchangeably.
\section{Conclusions and Future Work}
In this work, we introduce a benchmark to qualitatively evaluate different \seedword and \prompt approaches for extremely weakly supervised text classification. Through the benchmark, we raise awareness of the existence of \seedword approaches, that are strong competitors to the more well-known zero-shot prompting (with calibrations). We also experiment on the robustness of these two approaches, and show that \seedword are more tolerant to the given human guidance changes, however also being more selective to the pre-trained language models. We also analyzed the connections of \seedword and \prompt approaches through the lens of a few representative methods of the two approaches and showed that the methodologies are converging more recently.
Finally, we also include a study on clustering as a calibration technique that was independently proposed for both approaches
, and show that it can be a good performance booster.

We envision future work in two directions. The first one would be to understand the source of robustness difference and design a method that can take the best of both worlds (see Section~\ref{sec:implications}). 
The other would be to scale up the experiments and test if the conclusions still hold for larger pre-trained language models.


\section*{Limitations}

\noindent\textbf{Limitation of Model Scale}
The benchmark only included the evaluation of moderate-size language models and did not experiment on large language models. We justify our reasons in Section~\ref{sec:llm} and Appendix~\ref{sec:appendix_llm} and include an evaluation of ChatGPT in Appendix~\ref{sec:appendix_llm}, showing that even human feedback fine-tuned large language models is far from perfect on \xws-\tc. However, we acknowledge that the current state of extremely weak supervision would be better understood and assessed if complete evaluations on state-of-the-art large language models, such as Instruct-GPT~\cite{instructgpt}, PaLM~\cite{palm}, and ChatGPT exist. While we lack the computational resources to perform such an evaluation, we hope this work can stimulate interest in \xws-\tc and complete the study.

\noindent\textbf{Limitation of Text Classification}
\noindent Another limitation is the scope of Text Classification. While \prompt and \seedword methods have shown strong performances on text classification, this performance does not extend to other general classification tasks, such as natural language inference/entailment~\cite{npprompt}. 


\section*{Ethics Statement}
This paper establishes a benchmark for extremely weakly supervised text classification frameworks. We provide empirical results on various \seedword and \prompt methods, test their robustness, and analyze their connections. We give intuitions and insights on what method one should use for \xws-\tc in different circumstances. We believe that we are on the ethical side and do not find any ethical concerns in this work.

\bibliographystyle{acl_natbib}
\bibliography{anthology,custom}

\begin{thebibliography}{38}
\expandafter\ifx\csname natexlab\endcsname\relax\def\natexlab#1{#1}\fi

\bibitem[{Aharoni and Goldberg(2020)}]{unsup}
Roee Aharoni and Yoav Goldberg. 2020.
\newblock \href {https://doi.org/10.18653/v1/2020.acl-main.692} {Unsupervised
  domain clusters in pretrained language models}.
\newblock In \emph{Proceedings of the 58th Annual Meeting of the Association
  for Computational Linguistics, {ACL} 2020, Online, July 5-10, 2020}, pages
  7747--7763. Association for Computational Linguistics.

\bibitem[{Blei et~al.(2003)Blei, Ng, and Jordan}]{lda}
David~M. Blei, Andrew~Y. Ng, and Michael~I. Jordan. 2003.
\newblock \href {http://jmlr.org/papers/v3/blei03a.html} {Latent dirichlet
  allocation}.
\newblock \emph{J. Mach. Learn. Res.}, 3:993--1022.

\bibitem[{Brown et~al.(2020)Brown, Mann, Ryder, Subbiah, Kaplan, Dhariwal,
  Neelakantan, Shyam, Sastry, Askell, Agarwal, Herbert{-}Voss, Krueger,
  Henighan, Child, Ramesh, Ziegler, Wu, Winter, Hesse, Chen, Sigler, Litwin,
  Gray, Chess, Clark, Berner, McCandlish, Radford, Sutskever, and
  Amodei}]{gpt3}
Tom~B. Brown, Benjamin Mann, Nick Ryder, Melanie Subbiah, Jared Kaplan,
  Prafulla Dhariwal, Arvind Neelakantan, Pranav Shyam, Girish Sastry, Amanda
  Askell, Sandhini Agarwal, Ariel Herbert{-}Voss, Gretchen Krueger, Tom
  Henighan, Rewon Child, Aditya Ramesh, Daniel~M. Ziegler, Jeffrey Wu, Clemens
  Winter, Christopher Hesse, Mark Chen, Eric Sigler, Mateusz Litwin, Scott
  Gray, Benjamin Chess, Jack Clark, Christopher Berner, Sam McCandlish, Alec
  Radford, Ilya Sutskever, and Dario Amodei. 2020.
\newblock \href
  {https://proceedings.neurips.cc/paper/2020/hash/1457c0d6bfcb4967418bfb8ac142f64a-Abstract.html}
  {Language models are few-shot learners}.
\newblock In \emph{Advances in Neural Information Processing Systems 33: Annual
  Conference on Neural Information Processing Systems 2020, NeurIPS 2020,
  December 6-12, 2020, virtual}.

\bibitem[{Chowdhery et~al.(2022)Chowdhery, Narang, Devlin, Bosma, Mishra,
  Roberts, Barham, Chung, Sutton, Gehrmann, Schuh, Shi, Tsvyashchenko, Maynez,
  Rao, Barnes, Tay, Shazeer, Prabhakaran, Reif, Du, Hutchinson, Pope, Bradbury,
  Austin, Isard, Gur{-}Ari, Yin, Duke, Levskaya, Ghemawat, Dev, Michalewski,
  Garcia, Misra, Robinson, Fedus, Zhou, Ippolito, Luan, Lim, Zoph, Spiridonov,
  Sepassi, Dohan, Agrawal, Omernick, Dai, Pillai, Pellat, Lewkowycz, Moreira,
  Child, Polozov, Lee, Zhou, Wang, Saeta, Diaz, Firat, Catasta, Wei,
  Meier{-}Hellstern, Eck, Dean, Petrov, and Fiedel}]{palm}
Aakanksha Chowdhery, Sharan Narang, Jacob Devlin, Maarten Bosma, Gaurav Mishra,
  Adam Roberts, Paul Barham, Hyung~Won Chung, Charles Sutton, Sebastian
  Gehrmann, Parker Schuh, Kensen Shi, Sasha Tsvyashchenko, Joshua Maynez,
  Abhishek Rao, Parker Barnes, Yi~Tay, Noam Shazeer, Vinodkumar Prabhakaran,
  Emily Reif, Nan Du, Ben Hutchinson, Reiner Pope, James Bradbury, Jacob
  Austin, Michael Isard, Guy Gur{-}Ari, Pengcheng Yin, Toju Duke, Anselm
  Levskaya, Sanjay Ghemawat, Sunipa Dev, Henryk Michalewski, Xavier Garcia,
  Vedant Misra, Kevin Robinson, Liam Fedus, Denny Zhou, Daphne Ippolito, David
  Luan, Hyeontaek Lim, Barret Zoph, Alexander Spiridonov, Ryan Sepassi, David
  Dohan, Shivani Agrawal, Mark Omernick, Andrew~M. Dai,
  Thanumalayan~Sankaranarayana Pillai, Marie Pellat, Aitor Lewkowycz, Erica
  Moreira, Rewon Child, Oleksandr Polozov, Katherine Lee, Zongwei Zhou, Xuezhi
  Wang, Brennan Saeta, Mark Diaz, Orhan Firat, Michele Catasta, Jason Wei,
  Kathy Meier{-}Hellstern, Douglas Eck, Jeff Dean, Slav Petrov, and Noah
  Fiedel. 2022.
\newblock \href {https://doi.org/10.48550/arXiv.2204.02311} {Palm: Scaling
  language modeling with pathways}.
\newblock \emph{CoRR}, abs/2204.02311.

\bibitem[{Colombo et~al.(2022)Colombo, Noiry, Irurozki, and
  Cl{\'{e}}men{\c{c}}on}]{rankscore}
Pierre Colombo, Nathan Noiry, Ekhine Irurozki, and St{\'{e}}phan
  Cl{\'{e}}men{\c{c}}on. 2022.
\newblock \href {http://arxiv.org/abs/2202.03799} {What are the best systems?
  new perspectives on {NLP} benchmarking}.
\newblock \emph{CoRR}, abs/2202.03799.

\bibitem[{Devlin et~al.(2019)Devlin, Chang, Lee, and Toutanova}]{bert}
Jacob Devlin, Ming{-}Wei Chang, Kenton Lee, and Kristina Toutanova. 2019.
\newblock \href {https://doi.org/10.18653/v1/n19-1423} {{BERT:} pre-training of
  deep bidirectional transformers for language understanding}.
\newblock In \emph{Proceedings of the 2019 Conference of the North American
  Chapter of the Association for Computational Linguistics: Human Language
  Technologies, {NAACL-HLT} 2019, Minneapolis, MN, USA, June 2-7, 2019, Volume
  1 (Long and Short Papers)}, pages 4171--4186. Association for Computational
  Linguistics.

\bibitem[{Gao et~al.(2021)Gao, Fisch, and Chen}]{lmbff}
Tianyu Gao, Adam Fisch, and Danqi Chen. 2021.
\newblock \href {https://doi.org/10.18653/v1/2021.acl-long.295} {Making
  pre-trained language models better few-shot learners}.
\newblock In \emph{Proceedings of the 59th Annual Meeting of the Association
  for Computational Linguistics and the 11th International Joint Conference on
  Natural Language Processing, {ACL/IJCNLP} 2021, (Volume 1: Long Papers),
  Virtual Event, August 1-6, 2021}, pages 3816--3830. Association for
  Computational Linguistics.

\bibitem[{Gera et~al.(2022)Gera, Halfon, Shnarch, Perlitz, Ein{-}Dor, and
  Slonim}]{self_nli}
Ariel Gera, Alon Halfon, Eyal Shnarch, Yotam Perlitz, Liat Ein{-}Dor, and Noam
  Slonim. 2022.
\newblock \href {https://doi.org/10.48550/arXiv.2210.17541} {Zero-shot text
  classification with self-training}.
\newblock \emph{CoRR}, abs/2210.17541.

\bibitem[{Han et~al.(2022)Han, Hao, Dong, and Wei}]{protocal}
Zhixiong Han, Yaru Hao, Li~Dong, and Furu Wei. 2022.
\newblock \href {https://doi.org/10.48550/arXiv.2205.10183} {Prototypical
  calibration for few-shot learning of language models}.
\newblock \emph{CoRR}, abs/2205.10183.

\bibitem[{Holtzman et~al.(2021)Holtzman, West, Shwartz, Choi, and
  Zettlemoyer}]{dcpmi}
Ari Holtzman, Peter West, Vered Shwartz, Yejin Choi, and Luke Zettlemoyer.
  2021.
\newblock \href {https://doi.org/10.18653/v1/2021.emnlp-main.564} {Surface form
  competition: Why the highest probability answer isn't always right}.
\newblock In \emph{Proceedings of the 2021 Conference on Empirical Methods in
  Natural Language Processing, {EMNLP} 2021, Virtual Event / Punta Cana,
  Dominican Republic, 7-11 November, 2021}, pages 7038--7051. Association for
  Computational Linguistics.

\bibitem[{Hu et~al.(2022)Hu, Ding, Wang, Liu, Wang, Li, Wu, and Sun}]{KPT}
Shengding Hu, Ning Ding, Huadong Wang, Zhiyuan Liu, Jingang Wang, Juanzi Li,
  Wei Wu, and Maosong Sun. 2022.
\newblock \href {https://doi.org/10.18653/v1/2022.acl-long.158} {Knowledgeable
  prompt-tuning: Incorporating knowledge into prompt verbalizer for text
  classification}.
\newblock In \emph{Proceedings of the 60th Annual Meeting of the Association
  for Computational Linguistics (Volume 1: Long Papers), {ACL} 2022, Dublin,
  Ireland, May 22-27, 2022}, pages 2225--2240. Association for Computational
  Linguistics.

\bibitem[{Lang(1995)}]{20news}
Ken Lang. 1995.
\newblock Newsweeder: Learning to filter netnews.
\newblock In \emph{{ICML}}, pages 331--339. Morgan Kaufmann.

\bibitem[{Liang et~al.(2020)Liang, Yu, Jiang, Er, Wang, Zhao, and Zhang}]{bond}
Chen Liang, Yue Yu, Haoming Jiang, Siawpeng Er, Ruijia Wang, Tuo Zhao, and Chao
  Zhang. 2020.
\newblock \href {https://doi.org/10.1145/3394486.3403149} {{BOND:}
  bert-assisted open-domain named entity recognition with distant supervision}.
\newblock In \emph{{KDD} '20: The 26th {ACM} {SIGKDD} Conference on Knowledge
  Discovery and Data Mining, Virtual Event, CA, USA, August 23-27, 2020}, pages
  1054--1064. {ACM}.

\bibitem[{Liu et~al.(2019)Liu, Ott, Goyal, Du, Joshi, Chen, Levy, Lewis,
  Zettlemoyer, and Stoyanov}]{roberta}
Yinhan Liu, Myle Ott, Naman Goyal, Jingfei Du, Mandar Joshi, Danqi Chen, Omer
  Levy, Mike Lewis, Luke Zettlemoyer, and Veselin Stoyanov. 2019.
\newblock \href {http://arxiv.org/abs/1907.11692} {Roberta: {A} robustly
  optimized {BERT} pretraining approach}.
\newblock \emph{CoRR}, abs/1907.11692.

\bibitem[{Lu et~al.(2022)Lu, Bartolo, Moore, Riedel, and
  Stenetorp}]{promptorder}
Yao Lu, Max Bartolo, Alastair Moore, Sebastian Riedel, and Pontus Stenetorp.
  2022.
\newblock \href {https://doi.org/10.18653/v1/2022.acl-long.556} {Fantastically
  ordered prompts and where to find them: Overcoming few-shot prompt order
  sensitivity}.
\newblock In \emph{Proceedings of the 60th Annual Meeting of the Association
  for Computational Linguistics (Volume 1: Long Papers), {ACL} 2022, Dublin,
  Ireland, May 22-27, 2022}, pages 8086--8098. Association for Computational
  Linguistics.

\bibitem[{Ma et~al.(2023)Ma, Li, Lv, Zhu, Huang, and Hu}]{ProtoVerb}
Ting Ma, Mingming Li, Shangwen Lv, Fuqing Zhu, Longtao Huang, and Songlin Hu.
  2023.
\newblock \href {https://doi.org/10.1007/s10618-022-00851-2} {Conte:
  contextualized knowledge graph embedding for circular relations}.
\newblock \emph{Data Min. Knowl. Discov.}, 37(1):110--135.

\bibitem[{Maas et~al.(2011)Maas, Daly, Pham, Huang, Ng, and Potts}]{imdb}
Andrew~L. Maas, Raymond~E. Daly, Peter~T. Pham, Dan Huang, Andrew~Y. Ng, and
  Christopher Potts. 2011.
\newblock \href {http://www.aclweb.org/anthology/P11-1015} {Learning word
  vectors for sentiment analysis}.
\newblock In \emph{Proceedings of the 49th Annual Meeting of the Association
  for Computational Linguistics: Human Language Technologies}, pages 142--150,
  Portland, Oregon, USA. Association for Computational Linguistics.

\bibitem[{Mekala and Shang(2020)}]{conwea}
Dheeraj Mekala and Jingbo Shang. 2020.
\newblock \href {https://doi.org/10.18653/v1/2020.acl-main.30} {Contextualized
  weak supervision for text classification}.
\newblock In \emph{Proceedings of the 58th Annual Meeting of the Association
  for Computational Linguistics}, pages 323--333, Online. Association for
  Computational Linguistics.

\bibitem[{Meng et~al.(2020{\natexlab{a}})Meng, Huang, Wang, Wang, Zhang, Zhang,
  and Han}]{cate}
Yu~Meng, Jiaxin Huang, Guangyuan Wang, Zihan Wang, Chao Zhang, Yu~Zhang, and
  Jiawei Han. 2020{\natexlab{a}}.
\newblock Discriminative topic mining via category-name guided text embedding.
\newblock In \emph{{WWW}}, pages 2121--2132. {ACM} / {IW3C2}.

\bibitem[{Meng et~al.(2018)Meng, Shen, Zhang, and Han}]{westclass}
Yu~Meng, Jiaming Shen, Chao Zhang, and Jiawei Han. 2018.
\newblock \href {https://doi.org/10.1145/3269206.3271737} {Weakly-supervised
  neural text classification}.
\newblock In \emph{Proceedings of the 27th {ACM} International Conference on
  Information and Knowledge Management, {CIKM} 2018, Torino, Italy, October
  22-26, 2018}, pages 983--992. {ACM}.

\bibitem[{Meng et~al.(2020{\natexlab{b}})Meng, Zhang, Huang, Xiong, Ji, Zhang,
  and Han}]{lotclass}
Yu~Meng, Yunyi Zhang, Jiaxin Huang, Chenyan Xiong, Heng Ji, Chao Zhang, and
  Jiawei Han. 2020{\natexlab{b}}.
\newblock \href {https://doi.org/10.18653/v1/2020.emnlp-main.724} {Text
  classification using label names only: {A} language model self-training
  approach}.
\newblock In \emph{Proceedings of the 2020 Conference on Empirical Methods in
  Natural Language Processing, {EMNLP} 2020, Online, November 16-20, 2020},
  pages 9006--9017. Association for Computational Linguistics.

\bibitem[{Min et~al.(2022)Min, Lewis, Hajishirzi, and
  Zettlemoyer}]{noisy_channel}
Sewon Min, Mike Lewis, Hannaneh Hajishirzi, and Luke Zettlemoyer. 2022.
\newblock \href {https://doi.org/10.18653/v1/2022.acl-long.365} {Noisy channel
  language model prompting for few-shot text classification}.
\newblock In \emph{Proceedings of the 60th Annual Meeting of the Association
  for Computational Linguistics (Volume 1: Long Papers), {ACL} 2022, Dublin,
  Ireland, May 22-27, 2022}, pages 5316--5330. Association for Computational
  Linguistics.

\bibitem[{Ouyang et~al.(2022)Ouyang, Wu, Jiang, Almeida, Wainwright, Mishkin,
  Zhang, Agarwal, Slama, Ray, Schulman, Hilton, Kelton, Miller, Simens, Askell,
  Welinder, Christiano, Leike, and Lowe}]{instructgpt}
Long Ouyang, Jeff Wu, Xu~Jiang, Diogo Almeida, Carroll~L. Wainwright, Pamela
  Mishkin, Chong Zhang, Sandhini Agarwal, Katarina Slama, Alex Ray, John
  Schulman, Jacob Hilton, Fraser Kelton, Luke Miller, Maddie Simens, Amanda
  Askell, Peter Welinder, Paul~F. Christiano, Jan Leike, and Ryan Lowe. 2022.
\newblock \href {https://doi.org/10.48550/arXiv.2203.02155} {Training language
  models to follow instructions with human feedback}.
\newblock \emph{CoRR}, abs/2203.02155.

\bibitem[{Park and Lee(2022)}]{lime}
Seongmin Park and Jihwa Lee. 2022.
\newblock \href {https://aclanthology.org/2022.coling-1.91} {{LIME:}
  weakly-supervised text classification without seeds}.
\newblock In \emph{Proceedings of the 29th International Conference on
  Computational Linguistics, {COLING} 2022, Gyeongju, Republic of Korea,
  October 12-17, 2022}, pages 1083--1088. International Committee on
  Computational Linguistics.

\bibitem[{Perez et~al.(2021)Perez, Kiela, and Cho}]{truefewshot}
Ethan Perez, Douwe Kiela, and Kyunghyun Cho. 2021.
\newblock \href
  {https://proceedings.neurips.cc/paper/2021/hash/5c04925674920eb58467fb52ce4ef728-Abstract.html}
  {True few-shot learning with language models}.
\newblock In \emph{Advances in Neural Information Processing Systems 34: Annual
  Conference on Neural Information Processing Systems 2021, NeurIPS 2021,
  December 6-14, 2021, virtual}, pages 11054--11070.

\bibitem[{Roberts et~al.(2022)Roberts, Li, Huang, Adila, Schoenberg, Liu, Pick,
  Ma, Albarghouthi, and Sala}]{AutoWS-Bench-101}
Nicholas~Carl Roberts, Xintong Li, Tzu{-}Heng Huang, Dyah Adila, Spencer
  Schoenberg, Cheng{-}Yu Liu, Lauren Pick, Haotian Ma, Aws Albarghouthi, and
  Frederic Sala. 2022.
\newblock \href {https://doi.org/10.48550/arXiv.2208.14362} {Autows-bench-101:
  Benchmarking automated weak supervision with 100 labels}.
\newblock \emph{CoRR}, abs/2208.14362.

\bibitem[{Salton and Buckley(1988)}]{tfidf}
Gerard Salton and Chris Buckley. 1988.
\newblock \href {https://doi.org/10.1016/0306-4573(88)90021-0} {Term-weighting
  approaches in automatic text retrieval}.
\newblock \emph{Inf. Process. Manag.}, 24(5):513--523.

\bibitem[{Sanh et~al.(2022)Sanh, Webson, Raffel, Bach, Sutawika, Alyafeai,
  Chaffin, Stiegler, Raja, Dey, Bari, Xu, Thakker, Sharma, Szczechla, Kim,
  Chhablani, Nayak, Datta, Chang, Jiang, Wang, Manica, Shen, Yong, Pandey,
  Bawden, Wang, Neeraj, Rozen, Sharma, Santilli, F{\'{e}}vry, Fries, Teehan,
  Scao, Biderman, Gao, Wolf, and Rush}]{t0}
Victor Sanh, Albert Webson, Colin Raffel, Stephen Bach, Lintang Sutawika, Zaid
  Alyafeai, Antoine Chaffin, Arnaud Stiegler, Arun Raja, Manan Dey, M~Saiful
  Bari, Canwen Xu, Urmish Thakker, Shanya~Sharma Sharma, Eliza Szczechla,
  Taewoon Kim, Gunjan Chhablani, Nihal~V. Nayak, Debajyoti Datta, Jonathan
  Chang, Mike~Tian{-}Jian Jiang, Han Wang, Matteo Manica, Sheng Shen, Zheng~Xin
  Yong, Harshit Pandey, Rachel Bawden, Thomas Wang, Trishala Neeraj, Jos Rozen,
  Abheesht Sharma, Andrea Santilli, Thibault F{\'{e}}vry, Jason~Alan Fries,
  Ryan Teehan, Teven~Le Scao, Stella Biderman, Leo Gao, Thomas Wolf, and
  Alexander~M. Rush. 2022.
\newblock \href {https://openreview.net/forum?id=9Vrb9D0WI4} {Multitask
  prompted training enables zero-shot task generalization}.
\newblock In \emph{The Tenth International Conference on Learning
  Representations, {ICLR} 2022, Virtual Event, April 25-29, 2022}.
  OpenReview.net.

\bibitem[{Schick and Sch{\"{u}}tze(2021)}]{PET}
Timo Schick and Hinrich Sch{\"{u}}tze. 2021.
\newblock \href {https://doi.org/10.18653/v1/2021.eacl-main.20} {Exploiting
  cloze-questions for few-shot text classification and natural language
  inference}.
\newblock In \emph{Proceedings of the 16th Conference of the European Chapter
  of the Association for Computational Linguistics: Main Volume, {EACL} 2021,
  Online, April 19 - 23, 2021}, pages 255--269. Association for Computational
  Linguistics.

\bibitem[{Shang et~al.(2018)Shang, Liu, Jiang, Ren, Voss, and Han}]{autophrase}
Jingbo Shang, Jialu Liu, Meng Jiang, Xiang Ren, Clare~R. Voss, and Jiawei Han.
  2018.
\newblock \href {https://doi.org/10.1109/TKDE.2018.2812203} {Automated phrase
  mining from massive text corpora}.
\newblock \emph{{IEEE} Trans. Knowl. Data Eng.}, 30(10):1825--1837.

\bibitem[{Wang et~al.(2019)Wang, Singh, Michael, Hill, Levy, and Bowman}]{glue}
Alex Wang, Amanpreet Singh, Julian Michael, Felix Hill, Omer Levy, and
  Samuel~R. Bowman. 2019.
\newblock \href {https://openreview.net/forum?id=rJ4km2R5t7} {{GLUE:} {A}
  multi-task benchmark and analysis platform for natural language
  understanding}.
\newblock In \emph{7th International Conference on Learning Representations,
  {ICLR} 2019, New Orleans, LA, USA, May 6-9, 2019}. OpenReview.net.

\bibitem[{Wang et~al.(2021)Wang, Mekala, and Shang}]{xclass}
Zihan Wang, Dheeraj Mekala, and Jingbo Shang. 2021.
\newblock \href {https://doi.org/10.18653/v1/2021.naacl-main.242} {X-class:
  Text classification with extremely weak supervision}.
\newblock In \emph{Proceedings of the 2021 Conference of the North American
  Chapter of the Association for Computational Linguistics: Human Language
  Technologies, {NAACL-HLT} 2021, Online, June 6-11, 2021}, pages 3043--3053.
  Association for Computational Linguistics.

\bibitem[{Yin et~al.(2019)Yin, Hay, and Roth}]{nli}
Wenpeng Yin, Jamaal Hay, and Dan Roth. 2019.
\newblock \href {https://doi.org/10.18653/v1/D19-1404} {Benchmarking zero-shot
  text classification: Datasets, evaluation and entailment approach}.
\newblock In \emph{Proceedings of the 2019 Conference on Empirical Methods in
  Natural Language Processing and the 9th International Joint Conference on
  Natural Language Processing, {EMNLP-IJCNLP} 2019, Hong Kong, China, November
  3-7, 2019}, pages 3912--3921. Association for Computational Linguistics.

\bibitem[{Zhang et~al.(2021{\natexlab{a}})Zhang, Yu, Li, Wang, Yang, Yang, and
  Ratner}]{wrench}
Jieyu Zhang, Yue Yu, Yinghao Li, Yujing Wang, Yaming Yang, Mao Yang, and
  Alexander Ratner. 2021{\natexlab{a}}.
\newblock \href
  {https://datasets-benchmarks-proceedings.neurips.cc/paper/2021/hash/1c9ac0159c94d8d0cbedc973445af2da-Abstract-round2.html}
  {{WRENCH:} {A} comprehensive benchmark for weak supervision}.
\newblock In \emph{Proceedings of the Neural Information Processing Systems
  Track on Datasets and Benchmarks 1, NeurIPS Datasets and Benchmarks 2021,
  December 2021, virtual}.

\bibitem[{Zhang et~al.(2021{\natexlab{b}})Zhang, Ding, Xu, Liu, and
  Zhou}]{classkg}
Lu~Zhang, Jiandong Ding, Yi~Xu, Yingyao Liu, and Shuigeng Zhou.
  2021{\natexlab{b}}.
\newblock \href {https://doi.org/10.18653/v1/2021.emnlp-main.222}
  {Weakly-supervised text classification based on keyword graph}.
\newblock In \emph{Proceedings of the 2021 Conference on Empirical Methods in
  Natural Language Processing, {EMNLP} 2021, Virtual Event / Punta Cana,
  Dominican Republic, 7-11 November, 2021}, pages 2803--2813. Association for
  Computational Linguistics.

\bibitem[{Zhang et~al.(2015)Zhang, Zhao, and LeCun}]{charclass}
Xiang Zhang, Junbo~Jake Zhao, and Yann LeCun. 2015.
\newblock Character-level convolutional networks for text classification.
\newblock In \emph{{NIPS}}, pages 649--657.

\bibitem[{Zhao et~al.(2022)Zhao, Ouyang, Yu, Wu, and Li}]{npprompt}
Xuandong Zhao, Siqi Ouyang, Zhiguo Yu, Ming Wu, and Lei Li. 2022.
\newblock Pre-trained language models can be fully zero-shot learners.
\newblock \emph{arXiv preprint arXiv:2212.06950}.

\bibitem[{Zhao et~al.(2021)Zhao, Wallace, Feng, Klein, and Singh}]{contextcal}
Zihao Zhao, Eric Wallace, Shi Feng, Dan Klein, and Sameer Singh. 2021.
\newblock \href {http://proceedings.mlr.press/v139/zhao21c.html} {Calibrate
  before use: Improving few-shot performance of language models}.
\newblock In \emph{Proceedings of the 38th International Conference on Machine
  Learning, {ICML} 2021, 18-24 July 2021, Virtual Event}, volume 139 of
  \emph{Proceedings of Machine Learning Research}, pages 12697--12706. {PMLR}.

\end{thebibliography}

\clearpage\newpage
\appendix
\section{Other \seedword and \prompt methods}\label{sec:appendix_other}
\paragraph{More \seedword methods.}
There are also other \seedword methods that we will briefly describe here. WeSTClass~\cite{westclass} is one of the earlier weakly supervised methods that utilizes seed words to train a classifier by generating pseudo-documents instead of generating pseudo-labels. Conwea~\cite{conwea} explores the multi-sense of words and proposes to view seed words of different meanings as different words. Lime~\cite{lime} uses a fine-tuned model on a natural language inference dataset to suggest the seed words.

\paragraph{More \prompt methods.}
There are also other post/pre-processing techniques that we will briefly describe here. ContextualCal~\cite{contextcal} and PromptOrder~\cite{promptorder} work for in-context learning (in the few-shot scenario), and addresses the stability issue of the few-shot context in prompts. NosiyChannel~\cite{noisy_channel} considers the likelihood of generating the document based on the label, rather than generating the label based on the document.

\section{Dataset Sources}\label{sec:appendix_data}
The datasets are first introduced in the following papers:
\begin{itemize}[leftmargin=*,nosep]
    \item \textbf{IMDB}~\cite{imdb}.
    \item \textbf{Yelp-2, Yelp-5, AGNews,DBpedia}~\citet{charclass} 
    \item \textbf{20News, 20News-Fine} ~\citet{20news}\footnote{\url{http://qwone.com/~jason/20Newsgroups/}}
    \item \textbf{NYT-S, NYT-S-Fine,NYT, NYT-Loc}~\citet{cate}
\end{itemize}

\section{Detailed instructions and Label/Seed Words}\label{sec:appendix_instructions}
\begin{table*}[t]
    \centering
    \renewcommand\tabcolsep{3pt}
    \resizebox{\textwidth}{!}{
    \begin{tabular}{l | l | l}
    \toprule
    \textbf{Dataset} & \textbf{Instruction}                               & \textbf{Label Words/Seed Words} \\
    \midrule
    IMDB             & \shortstack[l]{review: <text>\\sentiment: <label>} & positive; negative \\ 
    \midrule
    Yelp-2           & \shortstack[l]{review: <text>\\sentiment: <label>} & positive; negative \\ 
    \midrule
    Yelp-5           & \shortstack[l]{review: <text>\\sentiment: <label>} & excellent; good; average; bad; awful \\ 
    \midrule
    AGNews      & \shortstack[l]{text: <text>\\topic: <label>} & \shortstack[l]{politics; sports; business; technology} \\ 
    \midrule
    20News      & \shortstack[l]{text: <text>\\topic: <label>} & \shortstack[l]{computer; sports; science; politics; religion} \\ 
    \midrule
    20News-Fine      & \shortstack[l]{text: <text>\\topic: <label>} & \shortstack[l]{atheism; graphics; Microsoft; IBM; Mac; motif; autos; motorcycles; baseball; \\hockey; encryption; electronics; medicine; space; Christian; guns; Arab} \\ 
    \midrule
    NYT-S      & \shortstack[l]{text: <text>\\topic: <label>} & \shortstack[l]{politics; art; business; science; sport} \\
    \midrule
    NYT-S-Fine      & \shortstack[l]{text: <text>\\topic: <label>} & \shortstack[l]{budget; gun; laws; gay; energy; environment; immigration; military; cosmos; \\insurance; stocks; bank; abortion; music; baseball; economy; television; golf; \\tennis; hockey; football; dance; movies; soccer; surveillance; basketball} \\
    \midrule
    NYT      & \shortstack[l]{text: <text>\\topic: <label>} & \shortstack[l]{business; politics; sports; health; education; estate; arts; science; technology} \\
    \midrule
    NYT-Loc      & \shortstack[l]{text: <text>\\location: <label>} & \shortstack[l]{America; Iraq; Japan; China; Britain; Russia; Germany; Canada; France; Italy} \\
    \midrule
    DBpedia      & \shortstack[l]{text: <text>\\topic: <label>} & \shortstack[l]{company; education; artist; athlete; politician; transportation; place; nature; \\village; species; plant; album; movie; book;
    } \\
    \bottomrule
    \end{tabular}
    }
    \caption{Instructions, Label words, and Seed Words.}
    \label{tab:instruction}
\end{table*}
We provide Table~\ref{tab:instruction} showing the instructions and label words used in the main experiment of the benchmark.


\section{Comparing Pre-trained Language Models}\label{sec:appendix_plm}
\begin{table*}[t]
    \centering
    \renewcommand\tabcolsep{3pt}
    \resizebox{\linewidth}{!}{
    \begin{tabular}{ll | ccc ccc ccc cc | cc}
    \toprule
    \textbf{Method} & \textbf{Model} & \textbf{IMDB} & \textbf{Yelp-2} & \textbf{Yelp-5} & \textbf{AGNews} & \textbf{20News} & \textbf{20News-Fine} & \textbf{NYT-S} & \textbf{NYT-S-Fine} & \textbf{NYT} & \textbf{NYT-Loc} & \textbf{DBpedia} & \textbf{Average} & \textbf{Rank Score} \\
    \midrule
    \multirow{6}{*}{Prompt}                             & GPT2-small   & 56.42 & 47.36 & 7.62 & 38.42 & 36.32 & 28.76 & 22.45 & 38.90 & 33.44 & 60.32 & 13.93 & 34.90 & 1 \\
    {}                                                  & BERT-base   & 42.16 & 35.48 & 7.59 & 68.89 & 50.35 & 3.78 & 49.94 & 39.96 & 37.88 & 38.49 & 17.71 & 35.67 & 0 \\
    {}                                                  & RoBERTa-base      & 40.51 & 54.01 & 15.27 & 66.94 & 46.87 & 12.45 & 33.27 & 19.80 & 38.88 & 43.92 & 28.60 & 36.41 & 2 \\
    \cmidrule{2-15}
    {}                                                  & GPT2-medium  & 35.80 & 33.57 & 25.87 & 69.36 & 55.16 & 46.03 & 54.08 & 46.14 & 24.92 & 79.00 & 24.52 & 44.95 & 4 \\ 
    {}                                                  & BERT-large   & 46.64 & 40.91 & 13.71 & 71.45 & 50.20 & 8.67 & 38.84 & 21.12 & 37.58 & 37.56 & 56.17 & 38.39 & 3 \\
    {}                                                  & RoBERTa-large      & 86.87 & 90.54 & 25.75 & 76.72 & 44.89 & 5.21 & 33.09 & 16.29 & 44.89 & 59.95 & 39.03 & 47.57 & 5 \\
    \bottomrule
    \end{tabular}
    }
    \caption{Performance of \prompt methods with different pre-trained language models. }
    \label{tab:prompt_performance}
\end{table*}
We are aware that a similar number of parameters in language models do not directly imply similar abilities.
We notice that the GPT-family LMs do tend to have a lower fine-tuning performance on natural language understanding tasks~\cite{glue} when compared with BERT/RoBERTa. 
However, we also notice that similar-sized GPT models do have a similar performance on zero-shot prompting as RoBERTa as observed in Table~\ref{tab:prompt_performance}. 
Since we are comparing under an \xws setting, instead of fully supervised fine-tuning, we believe it is fair to compare similar-size GPT models and RoBERTa models. 
We do acknowledge that BERT might be at a disadvantage since RoBERTa is better than BERT at both fully supervised fine-tuning~\cite{roberta} and zero-shot prompting (Table~\ref{tab:prompt_performance}). 
However, as we note in Section~\ref{sec:exp_model}, certain \seedword methods that work well on BERT might not be easily transferable to RoBERTa.

\section{Excluding Large Language Models}\label{sec:appendix_llm}
We did not include large language models in this benchmark. Here, we elaborate on two specific reasons.

From the design purpose of the benchmark, the focus of the benchmark is to understand the strengths of different \seedword and \prompt methods, which would be fruitful for moderate businesses or individual persons to make decisions on which method to use for \xws-\tc. Therefore, the analyses and comparisons on moderate-sized language models (100M - 300M parameters in the benchmark) would be more meaningful.

From a fair evaluation principle, all the models mentioned above are only developed for generative language models, which are not typically used for \seedword approaches. Using a more powerful language model for one approach would defeat the purpose of a fair comparison between models. Further, fine-tuned language models have already seen many classification tasks same as or very similar to the datasets in this benchmark. Therefore, it would be hard to access the true performance of the methods, as the similarity of the fine-tuned tasks to the evaluation tasks becomes another factor.

We also include an evaluation of ChatGPT on the benchmark. It is hard to fairly evaluate such a model, since 
(1) we do not know how it is trained and whether it saw the datasets in the benchmark, and
(2) there is no easy way to do large-scale evaluation.
We decide to evaluate it on the dataset NYT-S-Fine since we believe it is unlikely it is trained on such a fine-grained dataset. We pick 4 examples from each class resulting in total 104 examples. Since we can not retrieve the likelihoods, we embed the choice of classes in the prompt as follows: \verb|<instruction> <text> Answer:|, where \verb|<instruction>| is ``Choose exactly one of the following classes that best describes the text. Just give the class name as answer, no explanations, nothing more.'' followed by the list of all class names.

ChatGPT is able to suggest a single-word answer within the set of 26 class names in 91 out of 104 questions; we were able to correct 3 of the 13 out-of-scope answers since they do contain the correct class name. After the correction, ChatGPT is correct on 71 out of 104 questions, making it a model with 68.27\% prediction accuracy. The results of X-Class on the same 104 questions is 57.69\%. This indicates that while ChatGPT is performing pretty well, there is still much room to improve, given that it is using a much larger language model than X-Class is. 


\section{Method Implementations}\label{sec:appendix_implementations}
We use the public source implementation of different methods.

\noindent\textbf{X-Class} \url{https://github.com/ZihanWangKi/XClass}.

\noindent\textbf{LoTClass} \url{https://github.com/yumeng5/LOTClass}.

\noindent\textbf{ClassKG} \url{https://github.com/zhanglu-cst/ClassKG}.

\noindent\textbf{NPPrompt} \url{https://anonymous.4open.science/r/NPPrompt}.

\noindent\textbf{DCPMI} \url{https://github.com/peterwestuw/surface-form-competition}.

\noindent\textbf{ProtoCal} We implemented it ourselves.

\section{Computation Costs}\label{sec:appendix_computation}
We ran experiments on A6000 and A5000 GPUs. The total estimated GPU hours is 600.

\section{Full version of Table~\ref{tab:different_models}}\label{sec:appendix_differet_models_full}
\begin{table*}[th!]
    \centering
    \begin{tabular}{ll | cccc | cc}
    \toprule
    \textbf{Method} & \textbf{Model} & \textbf{Yelp-2} & \textbf{AGNews} & \textbf{NYT-S} &  \textbf{DBpedia} & \textbf{Average} & \textbf{Rank Score} \\
    \midrule
    \multicolumn{8}{c}{\prompt} \\
    \midrule\midrule
    \multirow{8}{*}{Prompt}                             & GPT2-small   & 47.36 & 38.42 & 22.45 & 13.93 & 30.54 & 1 \\
    {}                                                  & GPT2-medium  & 33.57 & 69.36 & 54.08 & 24.52 & 45.38 & 8 \\
    \cmidrule(lr){2-8}
    {}                                                  & BERT-base    & 35.58 & 68.89 & 49.94 & 17.71 & 43.04 & 7 \\
    {}                                                  & BERT-large   & 40.91 & 71.45 & 38.84 & 56.17 & 51.84 & 15 \\
    \cmidrule(lr){2-8}
    {}                                                  & RoBERTa-base & 54.01 & 66.94 & 33.27 & 28.60 & 45.71 & 6 \\
    {}                                                  & RoBERTa-large & 90.54 & 76.72 & 33.09 & 39.03 & 59.85 & 22 \\
    \cmidrule(lr){2-8}
    {}                                                  & BART-base    & 68.93 & 52.02 & 36.11 & 16.61 & 43.42 & 4 \\
    {}                                                  & BART-large   & 89.02 & 70.89 & 34.35 & 27.82 & 55.52 & 16 \\
    \midrule
    \multirow{8}{*}{\shortstack[l]{Prompt\\+ DCPMI}}    & GPT2-small   & 65.34 & 72.67 & 73.93 & 51.10 & 65.76 & 24 \\
    {}                                                  & GPT2-medium  & 87.00 & 74.13 & 79.80 & 57.30 & 74.56 & 31 \\
    \cmidrule(lr){2-8}
    {}                                                  & BERT-base    & 78.46 & 75.53 & 51.44 & 36.63 & 60.52 & 23 \\
    {}                                                  & BERT-large   & 78.02 & 64.38 & 21.09 & 60.02 & 55.88 & 14 \\
    \cmidrule(lr){2-8}
    {}                                                  & RoBERTa-base & 67.73 & 59.61 & 30.96 & 30.24 & 47.14 & 5\\
    {}                                                  & RoBERTa-large& 69.42 & 74.91 & 39.94 & 39.16 & 55.86 & 18 \\
    \cmidrule(lr){2-8}
    {}                                                  & BART-base    & 34.83 & 45.53 & 49.68 & 14.66 & 36.18 & 3 \\
    {}                                                  & BART-large   & 55.16 & 75.13 & 36.24 & 41.16 & 51.92 & 17 \\
    \midrule
    \multirow{8}{*}{\shortstack[l]{Prompt\\+ ProtoCal}} & GPT2-small   & 65.89 & 72.66 & 53.69 & 51.97 & 61.05 & 21 \\
    {}                                                  & GPT2-medium  & 88.60 & 75.26 & 51.97 & 64.46 & 70.07 & 30 \\
    \cmidrule(lr){2-8}
    {}                                                  & BERT-base    & 75.91 & 65.72 & 44.65 & 36.68 & 55.74 & 11 \\
    {}                                                  & BERT-large   & 78.18 & 66.45 & 57.51 & 78.52 & 70.16 & 25 \\
    \cmidrule(lr){2-8}
    {}                                                  & RoBERTa-base & 82.76 & 71.34 & 39.01 & 51.16 & 61.07 & 20 \\
    {}                                                  & RoBERTa-large& 92.13 & 78.95 & 43.29 & 49.97 & 66.09 & 28 \\
    \cmidrule(lr){2-8}
    {}                                                  & BART-base    & 86.78 & 52.94 & 47.51 & 23.51 & 52.68 & 10 \\
    {}                                                  & BART-large   & 92.18 & 73.89 & 50.73 & 50.83 & 66.91 & 27 \\
    \midrule
    \multicolumn{8}{c}{\seedword} \\
    \midrule\midrule
    \multirow{4}{*}{\shortstack[l]{X-Class}}            & BERT-base    & 85.44 & 81.81 & 91.94 & 89.50 & 87.17 & 37 \\
    {}                                                  & BERT-large   & 90.39 & 85.91 & 87.53 & 87.91 & 87.94 & 39 \\
    \cmidrule(lr){2-8}
    {}                                                  & RoBERTa-base & 55.06 & 32.66 & 61.17 & 91.85 & 60.18 & 19 \\
    {}                                                  & RoBERTa-large& 38.58 & 23.91 & 50.72 & 73.89 & 46.78 & 13 \\
    \midrule
    \multirow{4}{*}{\shortstack[l]{ClassKG}}            & BERT-base    & 92.21 & 88.10 & 84.12 & 94.75 & 89.80 & 40 \\
    {}                                                  & BERT-large   & 93.10 & 87.30 & 80.95 & 72.74 & 83.52 & 38 \\
    \cmidrule(lr){2-8}
    {}                                                  & RoBERTa-base & 79.04 & 88.84 & 82.98 & 96.89 & 86.94 & 36 \\
    {}                                                  & RoBERTa-large& 97.13 & 88.20 & 91.30 & 96.04 & 93.17 & 41 \\
    \midrule
    \multirow{4}{*}{\shortstack[l]{NPPrompt}}           & BERT-base    & 37.20 & 33.89 & 32.11 & 11.42 & 32.46 & 0 \\
    {}                                                  & BERT-large   & 37.20 & 33.89 & 13.49 & 41.20 & 31.45 & 2 \\
    \cmidrule(lr){2-8}
    {}                                                  & RoBERTa-base & 81.17 & 80.42 & 77.76 & 60.36 & 74.93 & 32 \\
    {}                                                  & RoBERTa-large& 93.58 & 83.62 & 77.93 & 47.11 & 75.56 & 33 \\
    \bottomrule
    \end{tabular}
    \caption{This is the full version of Table~\ref{tab:different_models}, that includes the performance of \prompt and \seedword methods when the choice of the pre-trained model is alternated. \prompt methods are evaluated on GPT2, BERT, BART, and RoBERTa, while \seedword methods are evaluated on BERT and RoBERTa.}
    \label{tab:different_models_full}
\end{table*}
We show Table~\ref{tab:different_models_full}, the detailed version of Table~\ref{tab:different_models} that includes performances on individual datasets.

\section{Full version of Table~\ref{tab:clustering}}\label{sec:appendix_clustering}
\begin{table*}[t]
    \centering
    \renewcommand\tabcolsep{3pt}
    \resizebox{\linewidth}{!}{
    \begin{tabular}{ll | ccc ccc ccc cc | cc}
    \toprule
    \textbf{Method} & \textbf{Model} & \textbf{IMDB} & \textbf{Yelp-2} & \textbf{Yelp-5} & \textbf{AGNews} & \textbf{20News} & \textbf{20News-Fine} & \textbf{NYT-S} & \textbf{NYT-S-Fine} & \textbf{NYT} & \textbf{NYT-Loc} & \textbf{DBpedia} & \textbf{Average} & \textbf{Rank Score} \\
    \midrule
    Prompt                             & GPT2-small   & 56.42 & 47.36 & 7.62 & 38.42 & 36.32 & 28.76 & 22.45 & 38.90 & 33.44 & 60.32 & 13.93 & 34.90 & 0 \\
    Prompt + clustering                & GPT2-small   & 70.35 & 65.89 & 23.77 & 72.66 & 58.62 & 36.77 & 53.69 & 29.82 & 55.15 & 65.80 & 51.97 & 53.14 & 1 \\
    \midrule
    Prompt + DCPMI                            & GPT2-small   & 70.13 & 65.34 & 23.01 & 72.67 & 61.64 & 37.45 & 73.93 & 63.19 & 55.20 & 70.40 & 51.10 & 58.55 & 2 \\
    Prompt + DCPMI + clustering                & GPT2-small   & 70.38 & 65.84 & 27.58 & 78.08 & 62.40 & 41.94 & 82.21 & 36.88 & 58.74 & 63.97 & 68.64 & 59.70 & 3 \\
    \midrule
    XClass (w/o clustering)            & BERT-base    & 73.79 & 83.49 & 27.48 & 72.05 & 74.09 & 55.35 & 85.76 & 55.93 & 68.57 & 82.37 & 62.48 & 67.40 & 6 \\
    XClass (w clustering)              & BERT-base    & 82.89 & 85.44 & 28.80 & 81.81 & 76.98 & 58.78 & 91.94 & 61.06 & 67.19 & 86.38 & 89.50 & 73.71 & 8 \\
    \midrule
    NPPrompt                           & RoBERTa-base & 85.19 & 81.17 & 14.20 & 80.42 & 68.92 & 48.64 & 77.76 & 55.23 & 64.46 & 53.85 & 60.36 & 62.75 & 4 \\
    NPPrompt + clustering              & RoBERTa-base & 84.84 & 82.99 & 14.48 & 83.12 & 70.42 & 50.44 & 91.84 & 44.10 & 62.22 & 54.17 & 71.32 & 64.54 & 5 \\
    \midrule
    ClassKG                            & BERT-base    & 88.08 & 92.21 & 32.33 & 88.10 & 81.72 & 52.29* & 84.12 & 49.59* & 60.79 & 92.81 & 94.75 & 74.25 & 7 \\
    ClassKG + clustering               & BERT-base    & 88.86 & 92.65 & 40.59 & 87.19 & 80.95 & 54.51* & 85.71 & 52.87* & 56.75 & 91.44 & 95.20 & 75.16 & 9 \\
    \bottomrule
    \end{tabular}
    }
    \caption{This is the full version of Table~\ref{tab:clustering} that contains the performance of \prompt and \seedword methods with and without the clustering post-processing. }
    \label{tab:clustering_full}
\end{table*}
We show Table~\ref{tab:clustering_full}, the detailed version of Table~\ref{tab:clustering} that includes performances on individual datasets.
\end{document}